\documentclass{article}
\usepackage{spconf,amsmath,graphicx}

\usepackage{flushend}
\usepackage{graphicx}
\title{A LARGE SCALE MULTI-VIEW RGBD VISUAL AFFORDANCE LEARNING DATASET}
\name{{Zeyad Khalifa$^{1,\dagger}$}, {Syed Afaq Ali Shah$^{1,2,\dagger}$}}
\address{$^{1}$School of Information Technology, Murdoch University, Perth, Australia \\ $^{2}$School of Science, Edith Cowan University, Perth, Australia 
\\$^{\dagger}$Both authors have equal contribution}

\begin{document}
%\ninept
%
\maketitle
\begin{abstract}
The physical and textural attributes of objects have been widely studied for recognition, detection and segmentation tasks in computer vision.~A number of datasets, such as large scale ImageNet, have been proposed for feature learning using data hungry deep neural networks and for hand-crafted feature extraction. To intelligently interact with objects, robots and intelligent machines need the ability to infer beyond the traditional physical/textural attributes, and understand/learn visual cues, called visual affordances, for affordance recognition, detection and segmentation. To date there is no publicly available large dataset for visual affordance understanding and learning. In this paper, we introduce a large scale multi-view RGBD visual affordance learning dataset, a benchmark of 47210 RGBD images from 37 object categories, annotated with 15 visual affordance categories. To the best of our knowledge, this is the first ever and the largest multi-view RGBD visual affordance learning dataset. We benchmark the proposed dataset for affordance segmentation and recognition tasks using popular Vision Transformer and Convolutional Neural Networks. Several state-of-the-art deep learning networks are evaluated each for affordance recognition and segmentation tasks. Our experimental results showcase the challenging nature of the dataset and present definite prospects for new and robust affordance learning algorithms. The dataset is publicly available at https://sites.google.com/view/afaqshah/dataset.
\end{abstract}
\section{Introduction}
\label{sec:intro}
Objects can be represented by various visual features such as shape, color, or physical attributes e.g., material, volume and weight. These attributes of objects are useful for object recognition, their segmentation or categorization into different classes, however they do not imply the potential actions that humans or intelligent machines can perform on objects while interacting with them. The ability of humans and machines to understand functional aspects of objects a.k.a object affordances has been studied for a long time \cite{gibson1979ecological}. In contrast to visual and physical attributes that mainly describe the object or its instance, affordance indicates functional interaction of humans with objects.

% \begin{figure}[t]
% \begin{center}
% %\fbox{\rule{0pt}{2in} \rule{0.9\linewidth}{0pt}}
%  \includegraphics[scale=.7]{samples.png}
% \end{center}
% \vspace{-2mm}
%  \caption{Few example images of bowl, flashlight and cereal box from our large scale multi-view visual affordance learning dataset (RGB images and affordance annotations).  Affordance labels such as wrap-grasp (green), containment (yellow), grasp (dark red), openable (blue) and rollable (gray) are represented with different colors. Figure best seen in color.} \label{fig:fig1}
%  \vspace{-2mm}
% \end{figure}
Object affordance understanding and learning is therefore crucial for intelligent machines such as autonomous robots to interact with objects in static as well as dynamic and complex scenes. Several applications require affordance learning e.g., predicting valid functionality of the objects and recognizing agent's actions in a given environment \cite{qi2017predicting}. 

One of the significant modalities for object affordance learning is through visual sensors, which capture visual affordances offered by an object \cite{hassanin2021visual}. Visual affordance has been recently explored with computer vision techniques. Luddecke et al., \cite{luddecke2017learning} used object parts segmentation to learn affordances. Castellini et al., \cite{castellini2011using} employed human hand poses during the interaction with objects to define affordances. Sun et al., \cite{sun2010learning} used object categorization to correctly infer the affordances. To achieve this, they developed an affordance model, which encoded the relationships among visual features, and learned categories and object affordances. A detailed review on the visual affordance understanding techniques and applications can be found in \cite{hassanin2021visual}.
\begin{figure*}[]
\begin{center}
%\fbox{\rule{0pt}{2in} \rule{0.9\linewidth}{0pt}}
 \includegraphics[scale=.45]{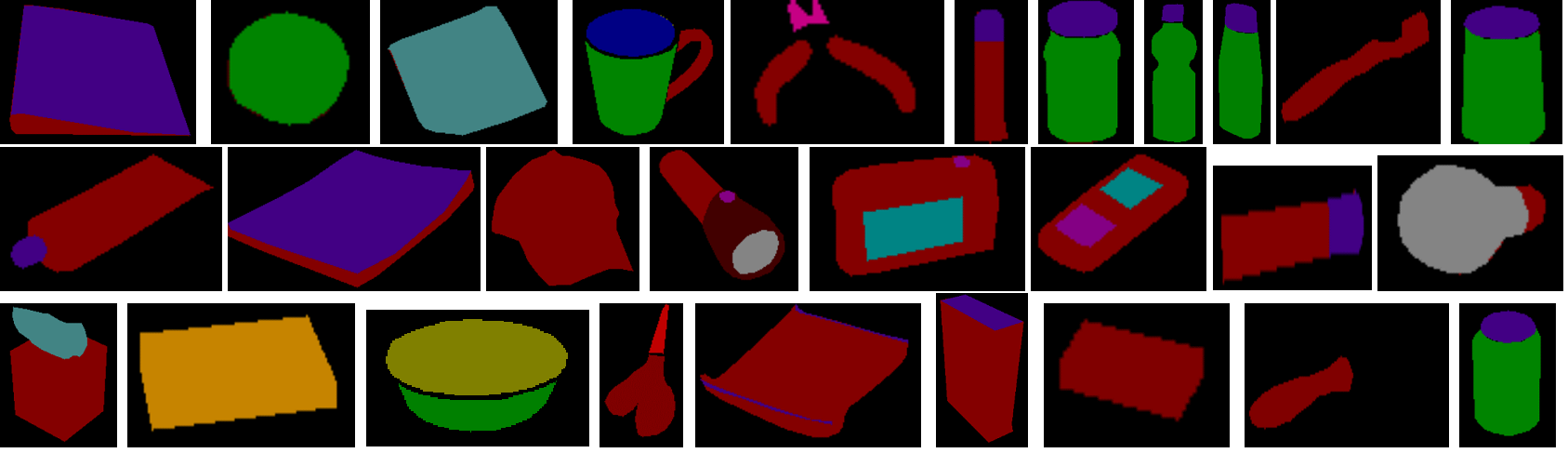}
\end{center}
\vspace{-2mm}
 \caption{Few example affordance annotations from our dataset. Objects include binder, ball, hand towel, coffee mug, plier, glue stick, food jar, water bottle, shampoo bottle, toothbrush,  food can, toothpaste, notebook, cap, flashlight, camera, calculator, glue stick, bulb, tissue box, sponge, bowl, scissors, noodle bag, cereal box, rubber eraser, comb, and soda can. Figure best seen in color.} \label{fig:fig1a}
 \vspace{-2mm}
\end{figure*}

Deep learning has received unprecedented performance for object recognition, detection and segmentation. %It is indeed dominating the field of vision and it has achieved remarkable improvements. 
In few cases, it has even achieved better performance than humans. The success of deep learning is yet to be seen for visual affordance learning. As pointed in \cite{hassanin2021visual}, one of the major bottlenecks is the small size of the existing affordance datasets. Deep learning algorithms need large annotated datasets for training and no such dataset is currently available for affordance learning \cite{hassanin2021visual}. In addition, the existing datasets are limited to a single view-point of objects. As a result, these datasets are not ideal for deep learning and do not pose significant challenges to modern computer vision techniques. 

To encourage research into visual affordance learning, particularly using deep neural networks, a large scale dataset is highly desired. Therefore, in this paper, we propose a large scale multi-view RGBD object affordance learning dataset. Fig.~\ref{fig:fig1a} shows example images from our object affordance dataset. Development of a large scale multi-view dataset is a challenging and subjective task. To handle the subjectivity problem, we take into account the affordance definitions from the existing research on visual affordance learning \cite{hassanin2021visual} and select possible affordance labels and interactions one can have with RGBD objects. We define 15 types of affordances over 37 indoor use object categories. In addition, we demonstrate that our dataset also enables benchmarking a wide range of tasks such as affordance recognition and affordance segmentation using deep learning. Several state-of-the-art deep neural networks are evaluated for affordance recognition and segmentation tasks. It is worth mentioning that affordance learning comes with the key underlying challenges. The same affordance can occur in very diverse scenarios/objects, e.g., a chair can afford sitting and a metal bar at the appropriate height can also afford sitting. In addition, it has one to many correspondences e.g., the same object can afford multiple functionalities. The proposed dataset overcomes these challenges.

The contributions of this paper can be summarised as:
\begin{itemize}
    \item We introduce a large scale multi-view RGBD affordance learning dataset, consisting of 47210 well-defined affordance information annotations covering 15 affordance classes and 37 object categories. To the best of our knowledge, this is the first large scale multi-view RGBD dataset for visual affordance learning. %Our dataset is still growing and we expect to add another 8000 RGBD annotated images soon.
    \item Our large scale dataset addresses the aforementioned challenges in affordance learning and helps bridge the gap, e.g., in our dataset a coffee mug and pitcher both can afford \textit{liquid containment}, similarly, hand towel and kleenex (tissue) afford \textit{dry} as one of the affordances. In addition, pitcher offers grasp, liquid containment, pourable and wrap-grasp i.e., multiple affordances.    
    \item We propose two affordance learning tasks including affordance recognition and segmentation to demonstrate the significance of the proposed dataset.
    \item Our experimental results demonstrate that the proposed dataset is valuable yet challenging for visual affordance learning tasks using deep neural networks.
\end{itemize}

%-------------------------------------------------------------------------
\section{Literature Review}
An affordance is the possibility of an action on an object or environment \cite{gibson1979ecological}. The idea of affordance learning in general is to identify the purpose, use, and ways to interact with an object, based on information gained from observing the object. It is an implementation of features that are learnt from various object classes, to predict how these objects may be used. Affordance learning is the core requirement in developing intelligent machines. In particular, visual affordance learning is the promising approach due to the diverse and multimodal information captured by the visual sensors. Due to recent advancements in visual sensors and data acquisition technology, a number of object affordance datasets have been developed for affordance recognition, segmentation and detection. In this section, we provide a brief overview of the existing affordance datasets.

Most of the visual affordance datasets consist of RGBD images and videos, except for 3D Affordance Net \cite{deng20213d}, which contains 3D point cloud data from PartNet \cite{mo2019partnet}. These datasets have been developed to benchmark specific task e.g., visual affordance segmentation and detection (UMD, CAD120, ADE-Affordance,  IIT-AFF and 3D Affordance Net). The other datasets benchmark object grasping and human interaction with objects for visual affordance understanding e.g., Binge Watching, HHOI, and COQE. These datasets feature 6 to 40 object categories, 1 to 15 affordance types (18 for point cloud based 3D Affordance Net, which is outside the scope of this paper due to a different modality) and 375 to 30k RGBD labelled images. The highest number of object categories are available in Extended NYUv2 \cite{roy2016multi}, however, this dataset consists of only 5 affordance types. On the other hand, ADE-Affordance features 15 affordance labels, however, it covers only 7 object categories. As also evident, the size of existing visual affordance datasets is very small e.g., UMD \cite{myers2015affordance} is the largest RGBD visual affordance dataset, which consists of 30k labelled images. These datasets are not big enough to train and test state-of-the-art deep learning networks. Deep learning based affordance algorithms need large annotated datasets, which are currently unavailable.

To address this challenge, we introduce a new large scale multi-view RGBD visual affordance learning dataset, which consists of 37 indoor object categories (sourced from a large scale hierarchical object dataset \cite{lai2011large}), 47210 RGBD labelled images and 15 affordance types, which are selectively inherited from a summary of existing works. Our dataset can be used to benchmark both visual affordance segmentation and recognition. In the following section, we provide details of our affordance learning dataset.

\section{Dataset Curation}
We present a large scale multi-view visual affordance learning dataset for affordance segmentation and recognition. To construct this dataset, raw RGBD images are collected from the Washington RGDB hierarchical multi-view object dataset \cite{lai2011large}, which covers 51 categories of common indoor objects. The Washington RGDB dataset was captured using a Kinect style 3D camera that records synchronized and aligned 640 x 480 RGB and depth images at 30 Hz. Objects were placed on a turntable and video sequences were captured for one whole rotation for each object. For each object, there are 3 video sequences, each recorded with the camera located at a different height relative to the turntable at approximately 30$^{o}$, 45$^{o}$, and 60$^{o}$ above the horizon. To avoid significant overlap between frames, we picked every 5th frame for annotation. From the available 51 object categories, we select those (37 object categories only) suitable for affordance learning and removed other objects e.g., fruits (apple, banana, pear etc.) and vegetables (mushroom, onion and potato etc). To ensure that our affordance dataset adds value to the research area we defined a set of affordance categories based on the existing work \cite{hassanin2021visual}. In total, 30 annotators were hired and trained for annotations. The average annotation time was around 3 minutes per image and each object category is annotated by two different annotators. All the annotations were verified by two experts. 
% \begin{figure*}[hbt]
% \begin{center}
% %\fbox{\rule{0pt}{2in} \rule{0.9\linewidth}{0pt}}
%  \includegraphics[scale=.55]{LabelMe.png}
% \end{center}
% \vspace{-2mm}
%  \caption{Annotation Tool (LabelMe). Left: Flashlight with four different affordance annotations. Right: Camera with three different affordances.} \label{fig:fig5aa}
%  \vspace{-2mm}
% \end{figure*}

\subsection{Affordance Labels and Annotation Tool}
From the list of affordance labels reported in \cite{hassanin2021visual}, we select those suitable for RGBD objects in \cite{lai2011large} and remove the irrelevant ones. In total, we filter out 12 types of affordances including, `grasp', `wrap grasp', `containment', `liquid-containment', `openable', `dry', `tip-push', `display', `illumination',  `cut', `pourable', and `rollable'. We also introduce three new affordances including `absorb', `grip' and `stapling' for sponge, plier and stapler, respectively. Next, we assign these affordances to each object category according to its functionality and affordances it offers to humans or robots.  

For affordance annotation, we used the popular LableMe graphical tool \cite{russell2008labelme}. The reason for choosing LabelMe is that it is a free open source tool, and it offers multiple features such as image annotation using polygon, circle, line, rectangle, and points. For our dataset, we used polygons for affordance annotations. We annotated RGB images using LabelMe, as these are visually easy to annotate compared to depth images, and then mapped those annotations on the corresponding depth images as the location of an object and affordance annotations are the same in both types of images.
% \begin{figure}[t]
% \begin{center}
% %\fbox{\rule{0pt}{2in} \rule{0.9\linewidth}{0pt}}
%  \includegraphics[scale=.7]{fig2.png}
% \end{center}
% \vspace{-2mm}
%  \caption{Dataset Annnotation. Left: Original RGB image. Center: Annotated image showing affordances such as display, tip-push and grasp. Right: Depth image. It is evident that the depth image is difficult to annotate compared to its RGB counterpart.} \label{fig:fig2}
%  \vspace{-2mm}
% \end{figure}

\subsection{Data Statistics}
 The dataset provides well-defined visual affordance annotations for 37 object categories with at most 4 affordance labels defined for each category. The dataset also contains 51304 affordance annotations from 15 affordance classes. Due to the multi-label nature of the affordance dataset, each object part could be labeled with multiple affordances e.g., pourable and liquid-containment. %Fig. \ref{fig:fig4} shows affordance annotations for few objects from our proposed dataset and Fig \ref{fig:fig3aa} shows the statistics of affordance labels in our dataset.
% \begin{figure}[hbt]
% \begin{center}
% %\fbox{\rule{0pt}{2in} \rule{0.9\linewidth}{0pt}}
%  \includegraphics[scale=.73]{aff_labels.png}
% \end{center}
% \vspace{-2mm}
%  \caption{Statistics of affordance labels in our dataset.} \label{fig:fig3aa}
%  \vspace{-2mm}
% \end{figure}

%-------------------------------------------------------------------------
\section{Benchmarking Results}
In this section, we benchmark two major tasks to demonstrate the effectiveness of our visual affordance learning dataset. These tasks include visual affordance segmentation and multi-label affordance recognition. %The former tackles the pixel-wise labelling of object parts for affordance segmentation and the latter deals with multi-label affordance categorization.

\subsection{Affordance Segmentation Results}
Affordance segmentation is a challenging problem. Given an object with not already known affordances, the affordance segmentation task aims to estimate the affordance types supported by the object and predict the probabilistic scores of affordance. For this task, we split the dataset into train, validation and test sets with a ratio of 80\%, 10\% and 10\%, respectively. The test set contains totally different shapes of objects, which the deep neural network has not seen during the training process.

\begin{table*}[htb]
  \centering
   \normalsize
   \vspace{2mm}
  \begin{tabular}{c | c |c | c }     \noalign{\hrule height 1pt}
Technique  &  mean IOU &  Frequency Weighted IOU &   Mean Accuracy\\    \noalign{\hrule height 1pt}
U-Net \cite{ronneberger2015u} & 39.25 & 40.21 & 49.10  \\ \hline
Mobile-Unet \cite{jing2020mobile}  & 38.83 & 39.76 & 52.00  \\ \hline
PSPNet-50 \cite{zhao2017pyramid}   & 40.12 & 42.75 & 54.61  \\ \hline
PSPNet-101 \cite{zhao2017pyramid}    & 40.88 & 43.82  & 55.93 \\ \hline
Segmenter \cite{strudel2021segmenter}   & 51.72 & 62.19  & 60.54  \\ \hline
HRViT \cite{gu2022multi}    & 52.50 & 53.72  & 63.38  \\ \hline
 AffordanceNet \cite{do2018affordancenet}    & 50.21 & 51.26  & 58.31  \\ \hline
%ST \cite{liu2021swin} & 56.87 & 67.09  & 65.75  \\ \hline
    \end{tabular}
     \caption{Visual affordance segmentation results. These experimental results demonstrate that our multi-view object affordance dataset is challenging for the state-of-the-art deep learning techniques.}\label{tab:table3}
%\vspace{5mm}
\end{table*}
We evaluate popular deep learning based segmentation architectures including U-Net \cite{ronneberger2015u}, Mobile-Unet \cite{jing2020mobile}, Pyramid Scene Parsing Network \cite{zhao2017pyramid}, AffordanceNet \cite{do2018affordancenet}; state-of-the-art Vision Transformers such as Segmenter (Transformer for Semantic Segmentation) \cite{strudel2021segmenter} and HRViT \cite{gu2022multi}. %, which are described here for completeness.
% \begin{table}[]
% \centering
% \begin{tabular}{c | c }
% \hline
% %\cline{1-2}
%    \textbf{Affordance} &  \textbf{IoU} \\ \hline
% grasp        & 85.65 \\ \hline
% wrap-grasp  & 75.66             \\ \hline
% containment  & 69.85  \\ \hline
% openable    & 66.27               \\ \hline
% tip-push     & 42.15                \\ \hline
% display      & 42.88              \\ \hline
% rollable    & 70.22               \\ \hline
% dry          & 50.25              \\ \hline
% liquid-containment   & 51.75         \\ \hline
% pourable  & 53.65 \\ \hline
% grip  & 55.22   \\ \hline
% absorb   & 61.29 \\ \hline
% cut & 41.85  \\ \hline
% stapling  & 42.75 \\ \hline
% illumination    & 43.84  \\ \hline 

% \end{tabular}
% \caption{IoU for each affordance type.}\label{tab:table3a}
% \end{table}

Table \ref{tab:table3} reports affordance segmentation results on our proposed affordance learning dataset. Since all of our images have background pixels, the chances of class imbalance are very high and the background pixels need to be weighed down compared to affordance classes. Frequency Weighted Intersection Over Union addresses this problem by taking a weighted mean based on the frequency of the class region in the dataset. All these models achieve a reasonable mean accuracy and frequency weighted IOU score, which indicates that the affordance estimation is a challenging task for the existing segmentation models. In contract, transformer based models achieve slightly better performance. These results demonstrate the challenging nature of the proposed dataset and also the robustness of ViTs. 

\subsection{Multi-label Visual Affordance Recognition}
In this section, we discuss and report results for visual affordance recognition task.  The affordance recognition task aims to categorize an image with the relevant set of affordance labels. To achieve this, an image is represented in the form of discriminative features and a classifier is used to predict affordance labels \cite{hassanin2021visual}. Affordance recognition is a multi-label problem, as opposed to object recognition, where each image only has one label. We therefore define and encode labels for each object category based on the affordance offered by an object. These encoded affordance labels are then used for the supervised training of deep neural networks. Note that the purpose of affordance recognition is not to classify an object, but to predict its affordance e.g., food can and food jar offer the same affordances i.e., wrap-grasp and openable, and can be used in lieu of each other. 
% \begin{table}[htb]
%   \centering
  
%   \normalsize
%    \vspace{2mm}
%   \begin{tabular}{c | c }     \noalign{\hrule height 1pt}

% Objects  &  Encoded Affordance Labels \\ \hline
%      \noalign{\hrule height 1pt}
% Ball             &  0 1 0 0 0 0 0 0 1 0 0 0 0 0 0 \\ \hline
% Binder           &  1 0 1 0 0 0 0 0 0 0 0 0 0 0 0 \\ \hline
% Bowl             &  0 1 0 1 0 0 0 0 0 0 0 0 0 0 0 \\ \hline
% Camera           &  1 0 0 0 0 1 0 0 0 0 0 0 0 0 0 \\ \hline
% Cap              &  1 0 0 0 0 0 0 0 0 0 0 0 0 0 0 \\ \hline
% Cell phone      &   1 0 0 0 1 1 0 0 0 0 0 0 0 0 0  \\ \hline
% Coffee mug      &  0 1 0 0 0 0 1 1 0 0 0 0 0 0 0\\ \hline
% Flashlight      &  1 0 0 0 0 1 0 0 1 1 0 0 0 0 0  \\ \hline
% Food bag        &  1 0 1 0 0 0 0 0 0 0 0 0 0 0 0 \\ \hline
% Pitcher          &  1 1 0 0 0 0 1 1 0 0 0 0 0 0 0 \\ \hline
% Sponge           &  1 0 0 0 0 0 0 0 0 0 0 0 0 1 0  \\ \hline
% Stapler          &  1 0 0 0 0 0 0 0 0 0 0 0 0 0 1 \\ \hline
%     \end{tabular}
%     \caption{Encoded affordance labels, shown as an example, for few object categories from our dataset. In each row, affordance labels appear in the following order: grasp, w-grasp, openable, containment, display, tip-push, liquid-containment, pourable, rollable, illumination, dry, cut, grip, absorb, stapling. 1 means the affordance type is offered by an object and 0 means the affordance is not available for the object category.}\label{tab:table4}
% %\vspace{5mm}
% \end{table}

We evaluate the performance of visual affordance recognition using the popular deep learning architectures including VGG-16 \cite{simonyan2014very}, ResNet-50 \cite{he2016deep} and InceptionV3 \cite{szegedy2015going}. For all the three models, a dropout layer of 0.5 is added after the last activation layer and before the last fully connected layer, where the network weights are initialised. We use Adam optimizer, learning rate is set to 0.0001, batch size and epochs are set to 64 and 20, respectively, for all the 3 models. 

We ran our experiments 5 folds for all the three models with random selection of training (70\%) and test (30\%) images. We do not perform any data augmentation and use raw images for the training of these models. Our affordance recognition results for all the three deep learning models are reported in Table \ref{tab:table5}, which reports average accuracy and the standard deviation. As can be noted these models achieve very good affordance recognition accuracy without any data augmentation, which demonstrate the effectiveness of our large scale multi-view RGBD dataset and its suitability for the training of deep neural networks.

\begin{table}[htb]
  \centering
   \normalsize
   \vspace{2mm}
  \begin{tabular}{c | c  }     \noalign{\hrule height 1pt}

Architecture  &  Average Accuracy (\%)\\    \noalign{\hrule height 1pt}
VGG-16 \cite{simonyan2014very}  &  85.17 $\pm$ 0.35 \\ \hline
ResNet-50 \cite{he2016deep} &  88.79 $\pm$ 1.55 \\ \hline
InceptionV3 \cite{szegedy2015going}  & 91.83 $\pm$ 0.35 \\ \hline
    \end{tabular}
     \caption{Visual Affordance Recognition Results. These experimental results demonstrate that our multi-view object affordance dataset is appropriate for the training and testing of deep neural networks without data augmentation.}\label{tab:table5}
%\vspace{5mm}
\end{table}
%------------------------------------------------------------------------
% \subsection{Implementation Details}
% All the deep learning architectures are implemented using TensorFlow and Keras libraries available in Python. The experiments were run on a Windows machine with 64 GB RAM, Core i7-9800X CPU @ 3.80GHz CPU and 12 GB NVIDIA Titan-V GPU.
%\vspace{-3mm}
\section{Conclusion}
\label{sec:conclusion}
In this paper, we have proposed a large scale multi-view RGBD visual affordance learning benchmark, which consists of 47210 RGBD images from 37 object categories, annotated with 15 visual affordance types. Using our dataset, we define two different affordance learning tasks and benchmarked six state-of-the-art deep networks including Vision Transformers, U-Net, Mobile-Unet, PSP-50, PSP-101 for affordance segmentation, and VGG-16, ResNet50, and Inception V3 for affordance recognition. Our results demonstrate that the proposed dataset is suitable for the training of deep neural networks for visual affordance learning tasks and without any data augmentation. The results also suggest that more robust techniques are required to perform affordance segmentation on our proposed benchmark. We believe that the proposed dataset will encourage the research community to focus on visual affordance learning research. 
% \vspace{-3mm}
% \section*{Acknowledgement}
% This research is supported by Edith Cowan University, Perth, Australia.
% References should be produced using the bibtex program from suitable
% BiBTeX files (here: strings, refs, manuals). The IEEEbib.bst bibliography
% style file from IEEE produces unsorted bibliography list.
% -------------------------------------------------------------------------
\bibliographystyle{IEEEbib}
\bibliography{refs}

\end{document}